\documentclass{article}

\usepackage[preprint]{neurips_2019}

\usepackage{color}

\usepackage[utf8]{inputenc} 
\usepackage[T1]{fontenc}    
\usepackage{hyperref}       
\usepackage{url}            
\usepackage{booktabs}       
\usepackage{amsfonts}       
\usepackage{nicefrac}       
\usepackage{microtype}      
\usepackage{footmisc}       
\usepackage{graphicx}
\usepackage{multirow}
\usepackage{amsmath}
\usepackage{algorithm}
\usepackage[noend]{algpseudocode}

\usepackage{amssymb}

\makeatletter
\def\BState{\State\hskip-\ALG@thistlm}
\makeatother

\title{Characterizing Bias in Classifiers using Generative Models}

\author{
  Daniel McDuff, Shuang Ma, Yale Song and Ashish Kapoor \\
  Microsoft\\
  Redmond,
  WA, USA \\
  \texttt{\{damcduff,v-mashua,yalesong,akapoor\}@microsoft.com} \\
}

\begin{document}

\maketitle

\begin{abstract}
Models that are learned from real-world data are often biased because the data used to train them is biased. This can propagate systemic human biases that exist and ultimately lead to inequitable treatment of people, especially minorities. To characterize bias in learned classifiers, existing approaches rely on human oracles labeling real-world examples to identify the ``blind spots'' of the classifiers; these are ultimately limited due to the human labor required and the finite nature of existing image examples. 
We propose a simulation-based approach for interrogating classifiers using generative adversarial models in a systematic manner. We incorporate a progressive conditional generative model for synthesizing photo-realistic facial images and Bayesian Optimization for an efficient interrogation of independent facial image classification systems. We show how this approach can be used to efficiently characterize racial and gender biases in commercial systems. 
\end{abstract}

\section{Introduction}
Models that are learned from found data (e.g., data scraped from the Internet) are often biased because these data sources are biased~\citep{torralba2011unbiased}. This can propagate systemic inequalities that exist in the real-world~\citep{caliskan2017semantics} and ultimately lead to unfair treatment of people. A model may perform poorly on populations that are minorities within the training set and present higher risks to them. For example, there is evidence of lower precision in pedestrian detection systems for people with darker skin tones (higher on the Fitzpatrick (\citeyear{fitzpatrick1988validity}) scale). This exposes one group to greater risk from self-driving/autonomous vehicles than another~\citep{wilson2019predictive}. Other studies have revealed systematic biases in facial classification systems~\citep{buolamwini2017gender,buolamwini2018gender}, with the error rate of gender classification up to seven times larger on women than men and poorer on people with darker skin tones. Another study found that face recognition systems misidentify people with darker skin tones, women, and younger people at higher error rates \citep{klare2012face}. To exacerbate the negative effects of inequitable performance, there is evidence that African Americans are subjected to higher rates of facial recognition searches~\citep{garvie2016perpetual}. Commercial facial classification APIs are already deployed in consumer-facing systems and are being used by law enforcement. The combination of greater exposure to algorithms and a reduced precision in the results for certain demographic groups deserves urgent attention.

\begin{figure*}[t]
\begin{center}
   \includegraphics[width=0.75\linewidth]{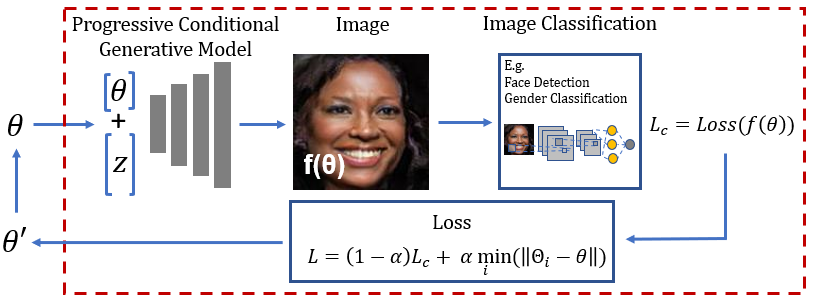}
\end{center}
   \caption{We propose a composite loss function, modeled as a Gaussian Process. It takes as input conditioning parameters for a validated progressive conditional generative model, and produces an image $f(\theta)$. This image is then used to interrogate an image classification system to compute a classification loss $L_c = Loss(f(\theta))$, where the loss is a binary classification loss (0 or 1), capturing whether the classifier performed accurately or poorly. This loss allows \textit{exploitation} of failures and is combined with a diversity loss to promote \textit{exploration} of the parameter space.}
\label{fig:intro}
\vspace{-0.5cm}
\end{figure*}

Many learned models exhibit bias as training datasets are limited in size and diversity.
Let us take several benchmark datasets as exemplars. Almost 50\% of the people featured in the widely used MS-CELEB-1M dataset~\citep{guo2016ms} are from North America (USA and Canada) and Western Europe (UK and Germany), and over 75\% are men.  The demographic make up of these countries is predominantly Caucasian/white.\footnote{http://data.un.org/\label{unstat}}  
Another dataset of faces, IMDB-WIKI~\citep{rothe2015dex}, features 59.3\% men and Americans are hugely over-represented (34.5\%). Another systematic profile~\citep{buolamwini2017gender} found that the IARPA Janus Benchmark A (IJB-A)~\citep{klare2015pushing} contained only 7.80\% of faces with skin types V or VI (on the Fitzpatrick skin type scale) and again featured over 75\% males. Sampling from the dataset listed here indiscriminately leads to a large proportion of images of males with lighter skin tones and upon training an image classifer often results in a biased system. Creating balanced datasets is a non-trivial task. Sourcing naturalistic images of a large number of different people is challenging. Furthermore, no matter how large the dataset is, it may still be difficult to find images that are distributed evenly across different demographic groups. 
Attempts have been made to improve facial classification by including gender and racial diversity. In one example, by Ryu et al. \citep{ryu2017improving}, results were improved by scraping images from the web and learning facial representations from a held-out dataset with a uniform distribution across race and gender intersections. 

Improving the performance of machine-learned classifiers is virtuous but there are other approaches to addressing concerns around bias. The concept of \textit{fairness through awareness}~\citep{dwork2012fairness} is the principle that in order to combat bias, we need to be aware of the biases and why they occur. In complex systems, such as deep neural networks, many of the ``unknowns'' are unknown and need to be identified \citep{lakkaraju2016discovering,lakkaraju2017identifying}. Identifying and characterizing ``unknowns'' in a model requires a combination of \textit{exploration} to identify regions of the model that contain failure modes and \textit{exploitation} to sample frequently from these region in order to characterize performance. Identifying failure modes is similar to finding adversarial examples for image classifiers~\citep{athalye2017synthesizing,tramer2017space}, a subject that is of increasing interest. 

One way of characterizing bias that holds promise is data simulation. Parameterized computer graphics simulators are one way of testing vision models~\citep{veeravasarapu2015model,veeravasarapu2015simulations,veeravasarapu2016model,vazquez2014virtual}. Generally, it has been proposed that graphics models be used for performance evaluation~\citep{haralick1992performance}.
Recently, McDuff et al.~\citeyear{mcduff2018identifying} illustrated how highly realistic simulations could be used to interrogate the performance of face detection systems. However, creating high fidelity 3D assets for simulating many different facial appearances (e.g., bone structures, facial attributes, skin tones etc.) is time consuming and expensive.

Generative adversarial networks (GANs)~\citep{GANs} are becoming increasingly popular for synthesizing data~\citep{shrivastava2017learning} and present an alternative, or complement, to graphical simulations. Generative models are trained to match the target distribution. Thus, once trained, a generative model can flexibly generate a large amount of diverse samples without the need for pre-built 3D assets. They can also be used to generate images with a set of desired properties by conditioning the model during the training stage and thus enabling generation of new samples in a controllable way at test time. Thus, GANs have been used for synthesizing images of faces at different ages~\citep{yang2017learning,choi2018stargan} or genders~\citep{dong2017unsupervised,choi2018stargan}. However, unlike parameterized models, statistical models (such as GANs) are fallible and might also have errors themselves. For example, the model may produce an image of a man even when conditioned on a woman. To use such a model for characterizing the performance of an independent classifier it is important to first characterize the error in the image generation itself. 

In this paper, we propose to use a characterized state-of-the-art progressive conditional generative model in order to test existing computer vision classification systems for bias, as shown in Figure~\ref{fig:intro}. In particular, we train a progressive conditional generative network that allows us to create high-fidelity images of new faces with different appearances by exploiting the underlying manifold.  We train the model on diverse image examples by sampling in a balanced manner from men and women from different countries. Then we characterize this generator using oracles (human judges) to identify any errors in the synthesis model. Using the conditioned synthesis model and a Bayesian search scheme we efficiently exploit and explore the parameterized space of faces, in order to find the failure cases of a set of existing commercial facial classification systems and identify biases. One advantage of this scheme is that we only need to train and characterize the performance of the generator once and can then evaluate many classifiers efficiently and systematically, with potentially many more variations of facial images than were used to train the generator.

The contributions of this paper are: (1) to present an approach for conditionally generating synthetic face images based on a curated dataset of people from different nations, (2) to show how synthetic data can be used to efficiently identify limits in existing facial classification systems, and (3) to propose a Bayesian Optimization based sampling procedure to identify these limits more efficiently. We release the nationality data, model and code to accompany the image data used in this paper (see the supplementary material).

\section{Related Work}

\textbf{Algorithmic Bias.}
There is wide concern about the equitable nature of machine learned systems. Algorithmic bias can exist for several reasons and the discovery or characterization of biases is non-trivial~\citep{hajian2016algorithmic}. Even if biases are not introduced maliciously they can result from explicit variables contained within a model or via variables that correlate with sensitive attributes - \textit{indirect discrimination}. Ideally we would minimize algorithmic bias or discrimination as much as possible, or prevent it entirely. However, this is challenging: First, algorithms can be released by third-parties who may not be acting in the public's best interest and not take the time or effort required to maximize the fairness of their models. Second, removing biases is technically challenging.  For example, balancing a dataset and removing correlates with sensitive variables is very difficult, especially when learning algorithms are data hungry and the largest, accessible data sources (e.g., the Internet) are biased~\citep{baeza2016data}. 

Tools are needed to help practitioners evaluate models, especially black-box models. Making algorithms more transparent and increasing accountability is another approach to increasing fairness~\citep{dwork2012fairness,lepri2018fair}. 
A significant study~\citep{buolamwini2018gender} highlighted that facial classification systems were not as accurate on faces with darker skin tones and on females. This paper led to improvements in the models behind these APIs being made~\citep{raji2019actionable}.  This illustrates how characterization of model biases can be used to advance the quality of machine learned systems. 

Biases often result from unknowns within a system. Methods have been proposed to help address the discovery of unknowns in predictive models~\citep{lakkaraju2016discovering,lakkaraju2017identifying}. In their work the search-space is partitioned into groups which can be given interpretable descriptions. Then an explore-exploit strategy is used to navigate through these groups systematically based on the feedback from an oracle (e.g., a human labeler). Bansal and Weld proposed a new class of utility models that rewarded how well the discovered $\textit{unknown unknowns}$ help explain a sample distribution of expected
queries \citep{bansal2018coverage}. Using human oracles is labor intensive and not scalable. We employ an explore-exploit strategy in our work, but rather than rely on a human oracle we use an image synthesis model. Using a conditioned model we provide a systematic way of interrogating a black box model by generating variations on the target images and repeatedly testing a classifier's performance. We incentivize the search algorithm to explore the parameter space of faces but also reward it for identifying failures and interrogating these regions of the space more frequently.

\textbf{Generative Adversarial Networks.}
Deep generative adversarial networks has enabled considerable improvements in image generation~\citep{goodfellow2014generative,zhang2017stackgan,xu2018attngan}. Conditional GANs~\citep{cGAN} allow for the addition of conditional variables, such that generation can be performed in a ``controllable'' way. The conditioning variables can take different forms (e.g. specific attributes or a raw image~\citep{choi2018stargan,yan2016attribute2image}.)
For facial images, this has been applied to control the gender~\citep{dong2017unsupervised}, age~\citep{yang2017learning,choi2018stargan}, hair color, skin tone and facial expressions~\citep{choi2018stargan} of generated faces. This allows for a level of systematic simulation via manifolds in the space of faces. Increasing the resolution of images synthesized using GANs is the focus of considerable research. Higher quality output images have been achieved by decomposing the generation process into different stages. The LR-GAN~\citep{Yang2017LRGANLR} decomposes the process by generating image foregrounds and backgrounds separately. 
StackGAN~\citep{zhang2017stackgan} decomposes generation stages into several steps each with greater resolution. PG-GAN~\citep{karras2018progressive} has shown impressive performance using a progressive training procedure starting from very low resolution (4$\times$4) and ending with high resolution images (1024$\times$1024). It can produce high fidelity images that are often tricky to distinguish from real photos.

In this paper, we employ a progressive conditional generative adversarial model for creating photo-realistic image examples with controllable ``gender'' and ``race''. These images are then used to interrogate independent image classification systems. We model the problem as a Gaussian process, sampling images from the model iteratively based on the performance of the classifiers, to efficiently discover blind-spots in the models. Empirically, we find that these examples can be used to identify biases in the image classification systems.

\section{Approach}
We propose to use a generative model to synthesize face images and then apply Bayesian optimization to efficiently generate images that have the highest likelihood of breaking a target classifier.

\textbf{Image Generation.}
To generate photo-realistic face images in a controllable way, we propose to adopt a progressively growing conditional GAN \citep{karras2018progressive,cGAN} architecture. This model is trained so as to condition the generator $G$ and discriminator $D$ on additional labels. The given condition $\theta$ could be any kinds of auxiliary information; here we use $\theta$ to specify both the race $r$ and gender $g$ of the subject in the image, i.e., $\theta=[r; g]$. During testing time, the trained $G$ should produce face images with the race and gender as specified by $\theta$.

We curated a dataset $\{x; \theta\}$ (described below), where $x$ is a face image and $\theta$ indicates the race $r$ and gender $g$ labels of $x$. To train the conditional generator, the input of the generator is a combination of a condition $\theta$ and a prior noise input $p_z(z)$; $z$ is a 100-D vector sampled from a unit normal distribution and $\theta$ is a one-hot vector that represents a unique combination of (race, gender) conditions. We concatenate $z$ and $\theta$ as the input to our model. $G$'s objective is defined by:
\begin{equation}
{\mathcal{L}_{G}} =  - {\mathbb{E}_{z,\theta} \big[\log D(G(z,\theta))\big]}
\end{equation}

The design of the discriminator $D$ is inspired by Thekumparampil et al.'s (\citeyear{NIPSRC-GAN}) Robust Conditional GAN model which proved successful at delivering robust results.
We train $D$ on two objectives: to discriminate whether the synthesized image is real or fake, and to classify the synthesized image into the correct class (e.g., race and gender). The training objective for $D$ is defined by:
\begin{equation}
{\mathcal{L}_{D}} = -{\mathbb{E}\big[\log D(x)\big]} - {\mathbb{E}_{z,\theta}\big[\log D(G(z,\theta))\big]} - {\mathbb{E}_{z,\theta}\big[\log C(G(z,\theta))\big]}
\end{equation}
where $C$ is an N-way classifier. Our full learning objective is:
\begin{equation}
\mathcal{L}_{adv} = \mathop {\min }\limits_{G} \mathop {\max }\limits_{D} {\mathcal{L}_{G}} + {\mathcal{L}_{D}}
\end{equation}

\begin{figure*}[t]
\begin{center}
   \includegraphics[width=\linewidth]{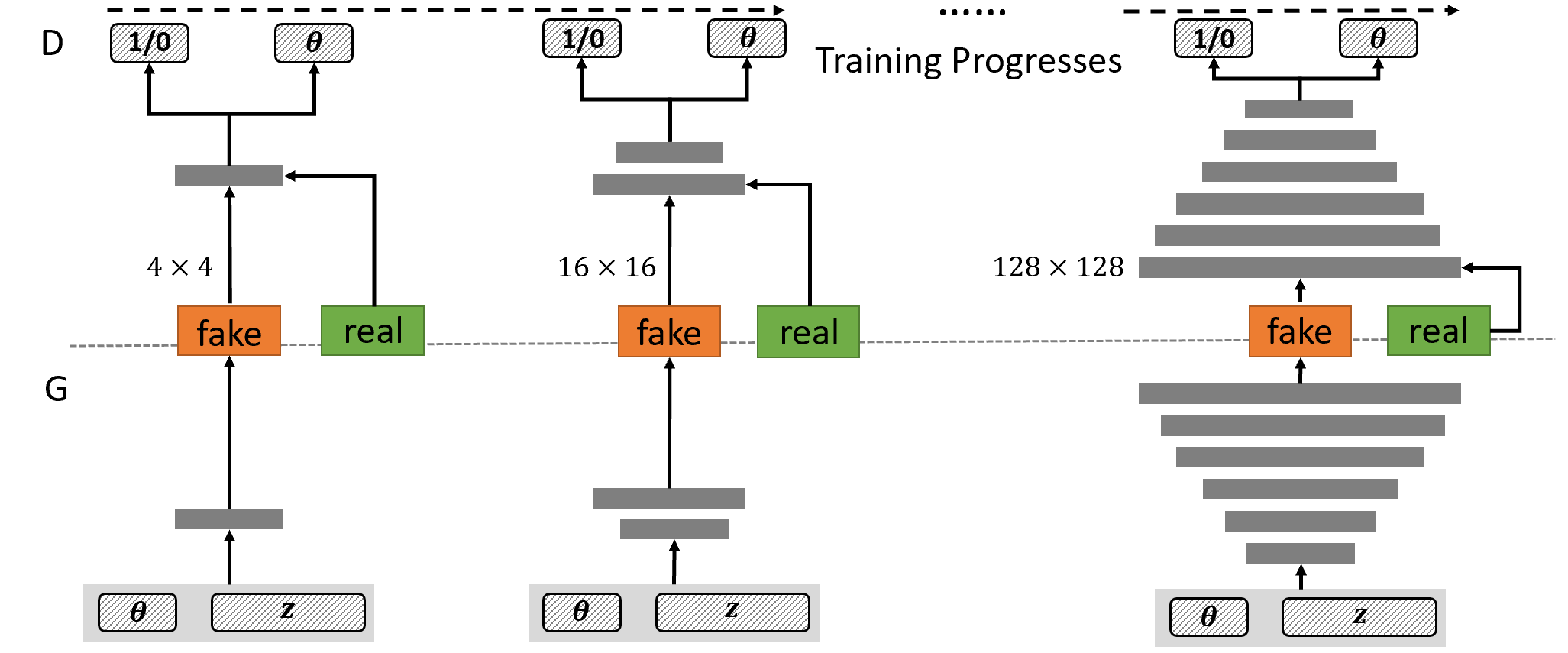}
\end{center}
   \caption{The training pipeline of our generative network. The input for generator $G$ is a joint hidden representation of combined latent vectors $z$ and labels $\theta$, where $\theta$ specifies the race and gender of the face. The Discriminator $D$ is used to discriminate the generated samples from real samples (indicated by $1/0$). At the same time, $D$ forces the generated samples to be classified into the appropriate corresponding classes (indicated by $\theta$). The network is trained progressively, i.e. from a resolution of 4$\times$4 pixels to 16$\times$16 pixels, and eventually to 128$\times$128 pixels (increasing by a factor of two).} 
\label{fig:architechture}
\vspace{-0.5cm}
\end{figure*}

We train the generator progressively~\citep{karras2018progressive} by increasing the image resolution by a power of two at each step, from 4$\times$4 pixels to 128$\times$128 pixels (see Figure~\ref{fig:architechture}). The real samples are downsampled into the corresponding resolution in each stage. The training code is included in supplementary material.

\textbf{Bayesian Optimization.}
Now that we have a systematically controllable image generation model, we propose to combine this with Bayesian Optimization \citep{brochu2010tutorial} to explore and exploit the space of parameters $\theta$ to find errors in the target classifier. We have $\theta$ as parameters that spawn an instance of a simulation $f(\theta)$ (e.g., a synthesized face image). This instance is fed into a target image classifier to check whether the system correctly identifies $f(\theta)$. Consequently, we can define a composite function $L_c = \mbox{\em{Loss}}(f(\theta))$, where $\mbox{\em{Loss}}$ is the classification loss and reflects if target classifier correctly handles the simulation instance generated when applying the parameters $\theta$. Carrying out Bayesian optimization with $L_c$ allows us to find $\theta$ that maximizes the loss, thus discovering parameters that are likely to break the classifier we are interrogating (i.e., \textit{exploitation}). However, we are not interested in just one instance but sets of diverse examples that would lead to misclassification. Consequently, we carry out a sequence of Bayesian optimization tasks, where each subsequent run considers an adaptive objective function that is conditioned on examples that were discovered in the previous round. Formally, in each round of Bayesian Optimization we maximize:
\begin{equation}
L = (1 - \alpha) L_c + \alpha \min_i{||\Theta_i - \theta||}.
\end{equation}
The above composite function is a convex combination of the misclassification cost with a term that encourages discovering new solutions $\theta$ that are diverse from the set of previously found examples $\Theta_i$. Specifically, the second term is the minimum euclidean distance of $\theta$ from the existing set and a high value of that term indicates that the example being considered is diverse from rest of the set. Intuitively, this term encourages \textit{exploration} and prioritizes sampling a diverse set of images. Figure~\ref{fig:intro} graphically describes such a composition. The sequence of Bayesian Optimizations find a diverse set of examples by first modeling the composite function $\textit{L}$ as a Gaussian Process (GP) \citep{rasmussen2004gaussian}. Modeling as a GP allows us to quantify uncertainty around the predictions, which in turn is used to efficiently explore the parameter space in order to identify the spots that satisfy the search criterion. In this work, we follow the recommendations in \cite{snoek2012practical}, and model the composite function via a GP with a Radial Basis Function (RBF) kernel, and use Expected Improvement (EI) as an acquisition function. The code is included in supplementary material.

\textbf{Data.}
We use the MS-CELEB-1M~\citep{guo2016ms} for our experimentation. This is a large image dataset containing 1M different people and approximately 100 million images.
To identify the nationalities of the people in the dataset we used the Google Search API and pulled biographic text associated with each person featured in the dataset. We then used the NLTK library to extract nationality and gender information from the biographies. Many nations have heterogeneous national and/or ethnic compositions and assuming that sampling from them at random would give consistent appearances is not well founded. Characterizing these differences is difficult, but necessary if we are to understand biases in vision classifiers. The United Nations (UN) notes that the ethnic and/or national groups of the population are dependent upon individual national circumstances and terms such as ``race'' and ``origin'' have many connotations. There is no internationally accepted criteria. Therefore, care must be taken in how we use these labels to generate images of different appearances. 

To help address this we used demographic data provided by the UN that gives the national and/or ethnic statistics for each country\footref{unstat} and then only sampled from countries with more homogeneous demographics. We selected four regions that have predominant and similar racial appearance groups. These group are Black (darker skin tones, Sub-Saharan African appearance), South Asian (darker skin tone, Caucasian appearance), Northeast Asian (moderate skin tone, East Asian appearance) and White (light skin tone, Caucasian appearance) and sampled from a set of countries to obtain images for each. We sampled 5,000 images (2,500 men and 2,500 women) from each region prioritizing higher resolution images (256$\times$256) and then lower resolution images (128$\times$128). The original raw images selected for training and the corresponding race and gender labels are included in the supplementary material. The nationality and gender labels for the complete MS-CELEB-1M will also be released.
Table~\ref{tab:dataset} shows the nations from which we sampled images and the corresponding appearance group. The number of people and images that were used in the final data are shown. It was not necessary to use all the images from every country to create a model for generating faces, and to obtain evenly distributed data over both gender and region we used this subset. Examples of the images produced by our trained model (described below) are also shown in the table. Higher resolution images can be found in the supplementary material.

\begin{table}
\small
\begin{center}
\begin{tabular}{rrcccccc}
\toprule
\multirow{2}{*}{Region} & \multirow{2}{*}{Country} & \multicolumn{2}{c}{People} & \multicolumn{2}{c}{Frames} & \multicolumn{2}{c}{Generated Images} \\
  & & M & W & M & W & M & W \\
\hline\hline
\multirow{5}{*}{Black}  &  &  &  &  &  & 
\multirow{1}{*}{\raisebox{-\totalheight}{\includegraphics[width=0.215\textwidth]{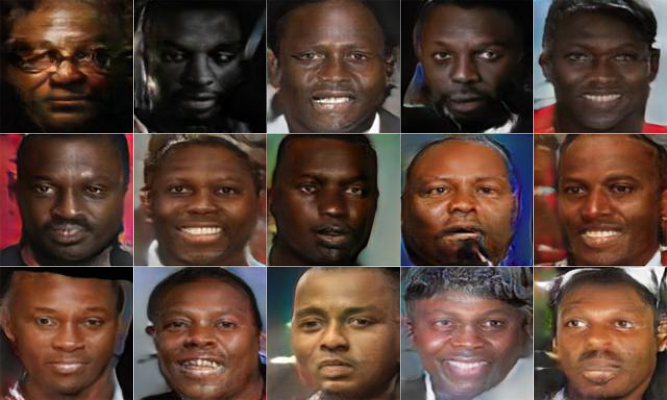}}} & \multirow{1}{*}{\raisebox{-\totalheight}{\includegraphics[width=0.215\textwidth]{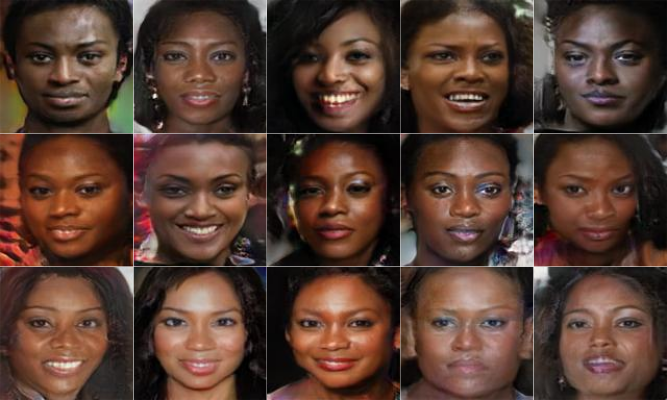}}} \\
& Nigerian & 81 & 28 & 768 & 467 & \\
 & Kenya & 11 & 5 & 91 & 49 \\
 & S. Africa & 136 & 102 & 1641 & 1984 & \\ \cline{3-4} \cline{5-6}
 &  \textbf{Total} & 228 & 135 & 2500 & 2500 &
 \\ 
 \multirow{5}{*}{S Asian}  &  & & & & &  \multirow{5}{*}{\raisebox{-\totalheight}{\includegraphics[width=0.215\textwidth]{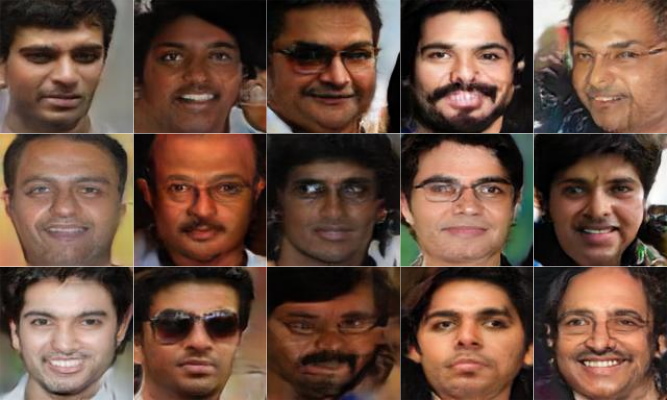}}} & \multirow{5}{*}{\raisebox{-\totalheight}{\includegraphics[width=0.215\textwidth]{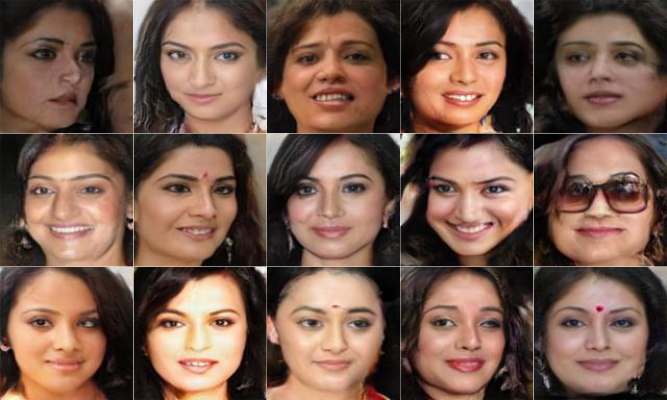}}} \\
 & India & 142 & 83 & 2108 & 2267  & \multirow{14}{*}{\raisebox{-\totalheight}{\includegraphics[width=0.215\textwidth]{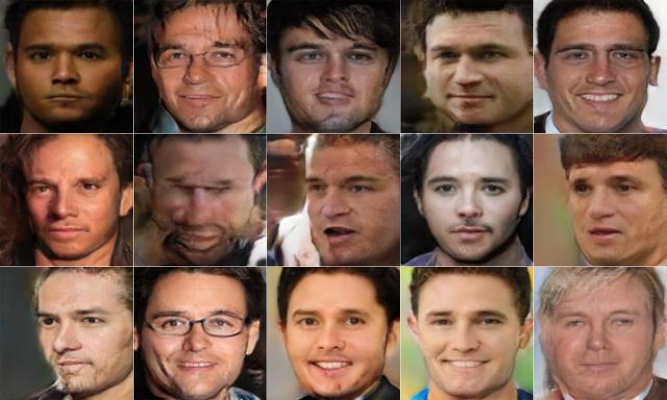}}} & \multirow{14}{*}{\raisebox{-\totalheight}{\includegraphics[width=0.215\textwidth]{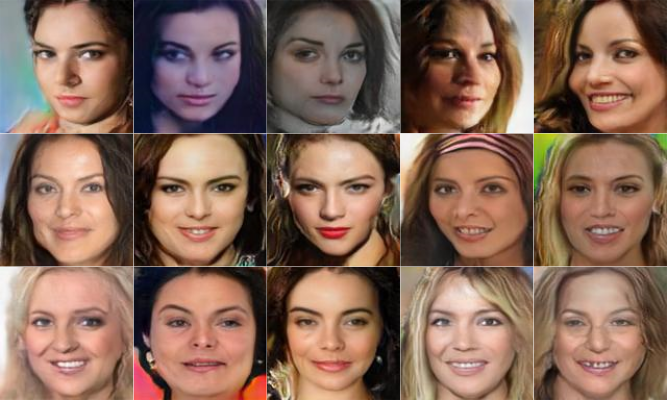}}} \\ 
& Sri Lanka & 1 & 2 & 11 & 7 &  \\
& Pakistan & 19 & 11 & 381 & 226  \\ \cline{3-4} \cline{5-6}
& \textbf{Total}  & 162 & 96 & 2500 & 2500 &  &  \\
\multirow{4}{*}{White}  &  & & & & & \\
  & Australia & 175 & 121 & 2500 & 2500 \\ \cline{3-4} \cline{5-6}
 & \textbf{Total} & 175 & 121 & 2500 & 2500 & & \\
\multirow{6}{*}{NE Asian} &  &  &  &  & &   \multirow{5}{*}{\raisebox{-\totalheight}{\includegraphics[width=0.215\textwidth]{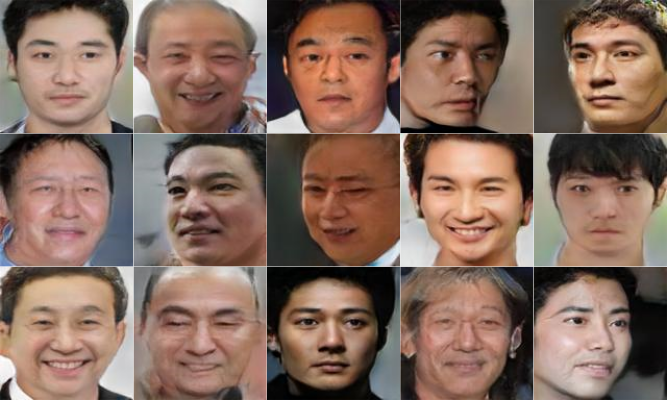}}} & \multirow{5}{*}{\raisebox{-\totalheight}{\includegraphics[width=0.215\textwidth]{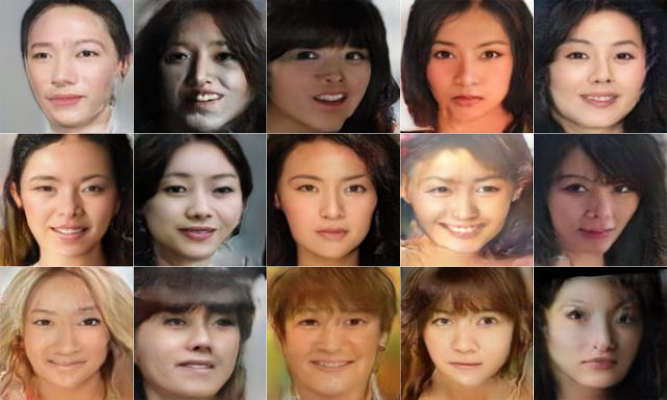}}} \\  
 & Japan & 105 & 89 & 930 & 1421 \\
 & China & 105 & 46 & 789 & 447 \\
 & S. Korea & 29 & 12 & 464 & 251 \\
 & Hong Kong & 36 & 28 & 317 & 381 \\ \cline{3-4} \cline{5-6}
 & \textbf{Total} & 275 & 175 & 2500 & 2500 &  &   \\ 
\bottomrule
\end{tabular}
\end{center}
\caption{The number of people and images we sampled from (by country and gender) to train our generation model.  Examples of generated faces for each race and gender. M = Men, W = Women.}
\label{tab:dataset}
\vspace{-0.4cm}
\end{table}

\section{Experiments and Results}

\textbf{Validation of Image Generation.}
Statistical generative models such as GANs are not perfect and may not always generate images that reflect the conditioned variables. Therefore, it is important to validate the performance of the GAN that we used at producing images that represent the specified conditions (race and gender) reliably. We generated a uniform sample of 50 images, at 128$\times$128 resolution, from each race and gender (total 50x4x2 = 400 images) and recruited five participants on MTurk to label the gender of the face in each image and the quality of the image (see Table~\ref{tab:dataset} for example images). The quality of the image was labeled on a scale of 0 (no face is identifiable) to 5 (the face is indistinguishable from a photograph). Of 400 images, the gender of only seven (1.75\%) images was classified differently by a majority of the labelers than the intended condition dictated. The mean quality rating of the images was 3.39 (SD=0.831) out of 5. There was no significant difference in quality between races or genders. In none of the images was a face considered unidentifiable. 

\textbf{Classifier Interrogation.}
Numerous companies offer services for face detection and gender detection from images (Microsoft, IBM, Amazon, SightEngine, Kairos, etc.).
We selected two of these commercial APIs (IBM and SightEngine) to interrogate in our experiments. These are exemplars and the specific APIs used here are not the focus of our paper. 
Each API accepts HTTP POST requests with URLs of the images or binary image data as a parameter within the request. If a face is detected they return JSON formatted data structures with the locations of the detected faces and a prediction of the gender of the face. Details of the APIs can be found in the supplementary material.

\begin{figure*}[t]
\begin{center}
   \includegraphics[width=\linewidth]{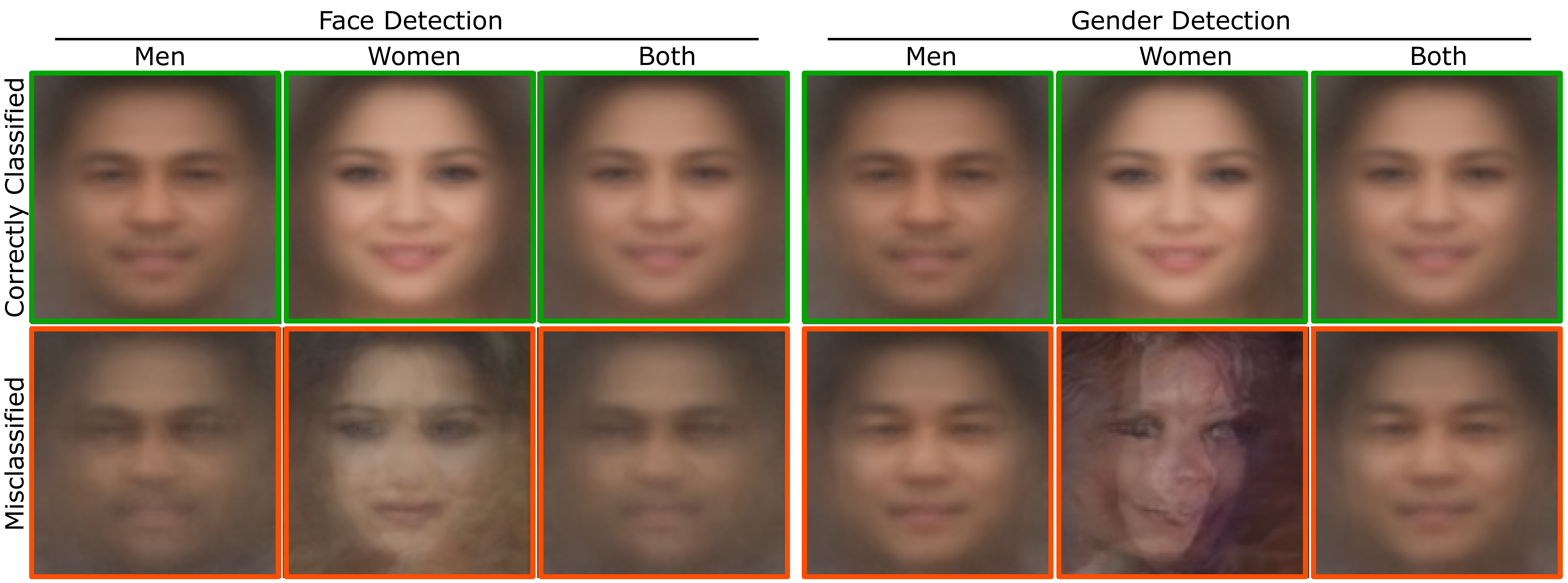}
\end{center}
   \caption{Mean faces for correct classifications and incorrect classifications. Notice how the skin tone for face detection failure cases is darker than for success cases. Women were very infrequently classified as men, thus the average face is not very clear.}
\label{fig:average_faces}
\vspace{-0.5cm}
\end{figure*}

We ran our sampling procedure for 400 iterations (i.e., we sampled 400 images at 128$\times$128 resolution) in each trial. Table~\ref{tab:results} shows the error rates (in \%) for face and gender detection. Figure~\ref{fig:average_faces} shows the mean face of images containing faces that were detected and not detected by the API. The skin tones illustrate that missed faces had darker skin tones and gender detection was considerably less accurate on people from NE Asia. We found men were more frequently misclassified as women than the other way around.

\begin{table}
\begin{center}
\begin{tabular}{cc|c|cccc|cc}
\toprule
API & Task & All  & Black & S Asian & NE Asian & White & Men & Women \\
\hline\hline
\multirow{2}{*}{IBM} & \multirow{1}{*}{Face Det.} & 8.05 & \textbf{16.9} & \textbf{7.63} & 3.96 & 3.8 & \textbf{11.3} & 2.27 \\ 
& \multirow{1}{*}{Gender Det.} & 8.26 & \textbf{9.00} & 2.13 & \textbf{20.0} & 1.87 & \textbf{15.8} & 0.27 \\
\bottomrule
\multirow{2}{*}{SE} & \multirow{1}{*}{Face Det.} & 0.13 & 0.00 & 0.00 & 0.53 & 0.00 & 0.21 & 0.00 \\ 
& \multirow{1}{*}{Gender Det.} & 2.84 & 3.39 & 0.74 & \textbf{5.85} & 1.38 & 5.14 & 0.00 \\
\bottomrule
\end{tabular}
\end{center}
\caption{Face detection and gender detection error rates (in percentage). SE = SightEngine. }
\label{tab:results}
\vspace{-0.5cm}
\end{table}

\begin{figure*}[t]
\begin{center}
   \includegraphics[width=\linewidth]{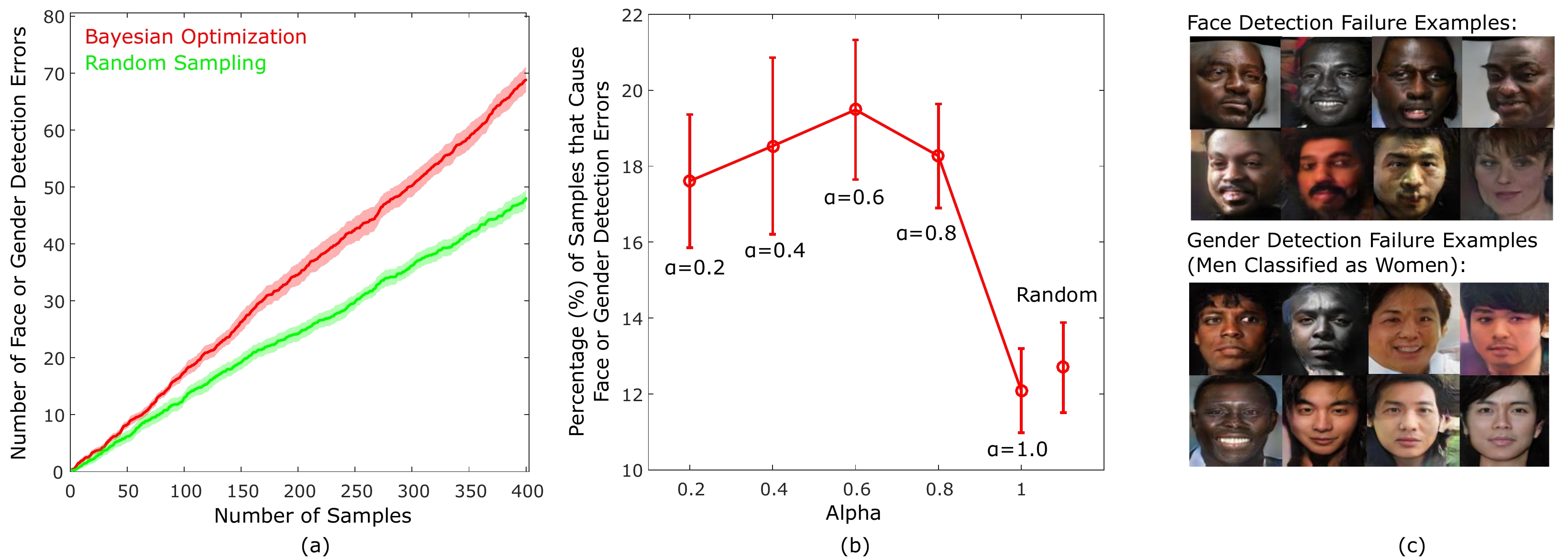}
\end{center}
   \caption{a) Sample efficiency of finding samples that were misclassified using random sampling and Bayesian Optimization with $\alpha$=1. Shaded regions reflect one standard deviation either side of the mean. b) Percentage of images that cause classifier failures (y-axis) as we vary the value of $\alpha$. Error bars reflect one standard deviation either side of the mean. c) Qualitative examples of failure cases.}
\label{fig:tradeoff}
\vspace{-0.5cm}
\end{figure*}

Next, we compare two approaches for searching our space of simulated faces for face detection and gender detection failure cases. For these analyses we used the IBM API as the target image classifier. First, we randomly sample parameters for generating face configurations and second we use Bayesian optimization. Again, we ran the sampling for 400 iterations in each case. In the case of Bayesian optimization, the image generation was updated dependant on the success or failure of classification of the previous image. This allows us to use an explore-exploit strategy and navigate through the facial appearance space efficiently using the feedback from the automated ``oracle''.  Figure~\ref{fig:tradeoff}(a) shows the sample efficiency of finding face detection and gender detection failures. Figure~\ref{fig:tradeoff}(b) shows the how the percentage of the errors found varies with the value of $\alpha$ and for random sampling.

\section{Discussion}

Bias in machine learning classifiers is problematic and often these biases may not be introduced intentionally. Regardless, biases can still propagate systemic inequalities that exist in the real-world. Yet, there are still few practical tools for helping researchers and developers mitigate bias and create well characterized classifiers. Adversarial training is a powerful tool for creating generative models that can produce highly realistic content. By using an adversarial training architecture we create a model that can be used to interrogate facial classification systems. We apply an optimal search algorithm that allows us to perform an efficient exploration of the space of faces to reveal biases from a smaller number of samples than via a brute force method. We tested this approach on face detection and gender detection tasks and interrogated commercial APIs to demonstrate its application. 

\textbf{Can our conditional GAN produce sufficiently high quality images to interrogate a classifier?} Our validation of our face GAN shows that the model is able to generate realistic face images that are reliably conditioned on race and gender. Human subjects showed very high agreement with the conditional labels for the generated images and the quality of the images were rated similarly across each race and gender. This suggests that our balanced data set and training procedure produced a model that can generate images reliably conditioned on race and gender and of suitably equivalent quality. Examples of the generated images can be seen in Table~\ref{tab:dataset} (high resolution images are available in the supplementary material). Very few of the images have very noticeable artifacts. 

\textbf{How do commercial APIs perform?}  Both of the commercial APIs we tested failed at significantly higher rates on images of people from the African and South Asian groups (see Table~\ref{tab:results}). For the IBM system the face detection error rate was more than four times as high on African faces as White and NE Asian faces. The error rates on Black and South Asian faces were the highest, suggesting that skin tone is a key variable here. Gender detection error rates were also high for African faces but unlike face detection the performance was worst for NE Asian faces. These results suggest that gender detection performance is not only impacted by skin tone but also other characteristics of appearance. Perhaps facial hair, or lack thereof, in NE Asian photographs, men with ``bangs'' and make-up (see Figure~\ref{fig:tradeoff}c and supplementary material for examples of images that resulted in errors.)

\textbf{Can we efficiently sample images to find errors?} Errors are typically sparse (many APIs have a global error rate of less than 10\%) and therefore simply randomly sampling images in order to identify biases is far from efficient. Using our optimization scheme we are able to identify an equivalent number of errors in significantly fewer samples (see Figure~\ref{fig:tradeoff}a). The results show that we are able to identify almost 50\% more failure cases using the Bayesian sampling scheme than without. In some senses our approach can be thought of as a way of efficiently identifying adversarial examples.  

\textbf{Trading off exploitation and exploration?} In our sampling procedure we have an explicit trade-off, using $\alpha$, between exploration of the underlying manifold of face images and exploitation to find the highest number of errors (see Figure~\ref{fig:tradeoff}(b)). With little exploration there is a danger that the sampling will find a local minima and continue sampling from a single region.  Our results show that with $\alpha$ equal 0.6 we maximize the number of errors found. This is empirical evidence that exploration and exploration are both important. Otherwise there is risk that one might miss regions that have frequent failure cases. As the parameter space of $\theta$ grows in dimensionality our sampling procedure will become even more favorable compared to naive methods.

\section{Conclusions}

We have presented an approach applying a conditional progressive generative model for creating photo-realistic synthetic images that can be used to interrogate facial classifiers. We test commercial image classification application programming interfaces and find evidence of systematic biases in their performance. A Bayesian search algorithm allows for efficient search and characterization of these biases. Biases in vision-based systems are of wide concern especially as these system become widely deployed by industry and governments. Generative models are a practical tool that can be used to characterize the performance of these systems. We hope that this work can help increase the prevalence of rigorous benchmarking of commercial classifier in the future.

\bibliographystyle{bib_style}
\bibliography{references}

\end{document}